\documentclass[letterpaper, 10 pt, conference]{ieeeconf}  

\IEEEoverridecommandlockouts                              

\usepackage{cite}
\usepackage{booktabs}
\usepackage{amsmath,amssymb,amsfonts}
\usepackage{algorithmic}
\usepackage{graphicx}
\usepackage{textcomp}
\usepackage{xcolor}
\usepackage{multirow}
\usepackage{multicol}
\usepackage{subcaption}
 \usepackage[caption=false,font=footnotesize]{subfig}

 \usepackage{url}

\def\BibTeX{{\rm B\kern-.05em{\sc i\kern-.025em b}\kern-.08em
    T\kern-.1667em\lower.7ex\hbox{E}\kern-.125emX}}

\title{\LARGE \bf
Preparation of Papers for IEEE Sponsored Conferences \& Symposia*
}

\author{Salim Khazem$^{1}$
\thanks{*This work was not supported by Talan Company}
\thanks{$^{1}$Salim Khazem is with Talan Research Center, Paris, France
        {\tt\small salim.talan@talan.com}}%
}

\begin{document}

\title{Multi-Scale Visual Prompting for Lightweight Small-Image Classification}


\maketitle

\begin{abstract}
Visual prompting has recently emerged as an efficient strategy to adapt vision models using lightweight, learnable parameters injected into the input space. However, prior work mainly targets large Vision Transformers and high-resolution datasets such as ImageNet. In contrast, small-image benchmarks like MNIST, Fashion-MNIST, and CIFAR-10 remain widely used in education, prototyping, and research, yet have received little attention in the context of prompting. In this paper, we introduce \textbf{Multi-Scale Visual Prompting (MSVP)}, a simple and generic module that learns a set of global, mid-scale, and local prompt maps fused with the input image via a lightweight $1 \times 1$ convolution. MSVP is backbone-agnostic, adds less than $0.02\%$ parameters, and significantly improves performance across CNN and Vision Transformer backbones.

We provide a unified benchmark on MNIST, Fashion-MNIST, and CIFAR-10 using a simple CNN, ResNet-18, and a small Vision Transformer. Our method yields consistent improvements with negligible computational overhead. We further include ablations on prompt scales, fusion strategies, and backbone architectures, along with qualitative analyzes using prompt visualizations and Grad-CAM. Our results demonstrate that multi-scale prompting provides an effective inductive bias even on low-resolution images. The code is available at \url{https://github.com/salimkhazem/msvp.git}

\end{abstract}


\section{Introduction}
Small-image benchmarks such as MNIST~\cite{lecun1998mnist}, Fashion-MNIST~\cite{xiao2017fashionmnist} and CIFAR-10~\cite{krizhevsky2009cifar} have played a central role in the history of deep learning for computer vision. Despite their apparent simplicity, they continue to be used extensively for algorithmic prototyping, educational purposes, and controlled method comparisons. These datasets expose practical challenges related to generalization, calibration, and robustness, especially when models are constrained to be small and computationally efficient.

At the same time, the last few years have witnessed a rapid shift towards \emph{parameter-efficient tuning} methods such as LoRA~\cite{hu2022lora}, adapters~\cite{houlsby2019adapter}, and prompt tuning~\cite{lester2021power,li2021prefix}. Instead of fully fine-tuning all parameters of a large pre-trained model, many approaches propose to adapt only a small subset of additional parameters. In natural language processing, soft prompts and low-rank adapters have become standard tools. In computer vision, \emph{Visual Prompt Tuning} (VPT)~\cite{jia2022vpt} and its extensions~\cite{yoo2023improving, wang2024revisiting, jin2025lorvp, kim2024fewprompts} show that Vision Transformers (ViTs)~\cite{dosovitskiy2020image} can be adapted to new tasks by introducing learnable prompt tokens at the input or intermediate layers, while freezing the majority of the backbone.

Most of this work, however, focuses on high-resolution tasks (e.g., ImageNet~\cite{krizhevsky2012imagenet}) and large pre-trained ViTs. In such setups, visual prompting serves primarily as a mechanism for task adaptation and transfer learning. In contrast, small-image datasets are often tackled with models trained \emph{from scratch}, and the role of prompting in this regime is largely unexplored. This raises a natural question:

\begin{quote}
\emph{Can we leverage the idea of visual prompting to obtain systematic gains on low-resolution benchmarks and small backbones, even when no large-scale pre-training is used?}
\end{quote}

In this paper, we answer this question in the affirmative by proposing \textbf{Multi-Scale Visual Prompting (MSVP)}. Instead of introducing abstract prompt tokens in a sequence of patch embeddings, MSVP operates directly in \emph{pixel space}. It maintains a small set of learnable prompt maps at different spatial scales (global, mid-scale, local), upsamples them to the image resolution, and fuses them with the original image via a $1 \times 1$ convolution. The result is a \emph{prompt-augmented image} that can be fed into any backbone architecture.

\begin{figure}[h]
\centering
\includegraphics[width=0.48\textwidth]{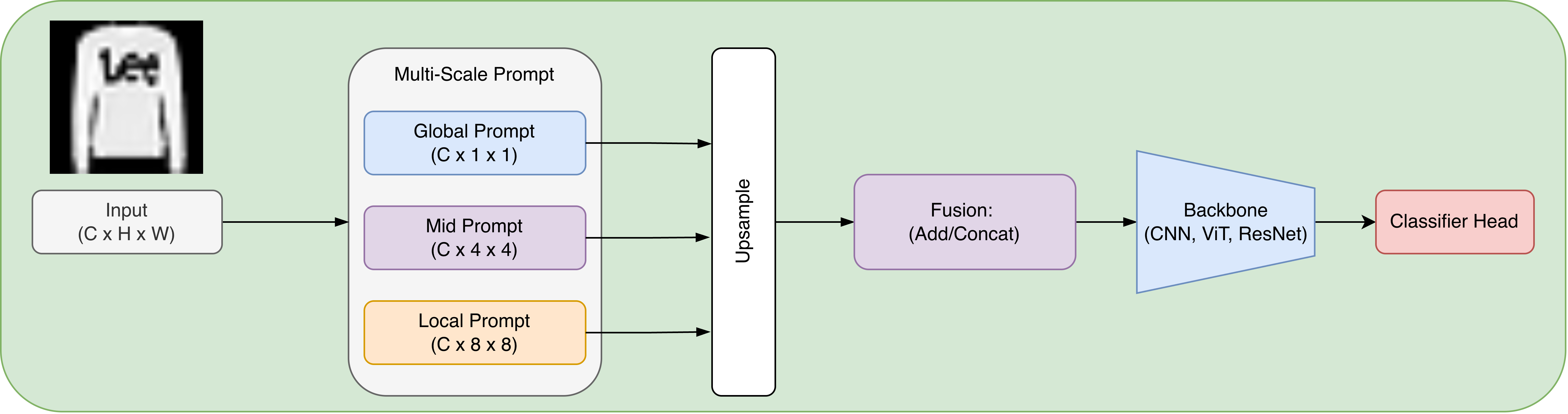}
\caption{Multi-Scale Visual Prompting architecture. Three learnable prompts 
at different spatial scales (global $1\times1$, mid $4\times4$, local $8\times8$) 
are upsampled to input resolution and fused via element-wise addition.}
\label{fig:architecture}
\end{figure}

Our main hypothesis is that these multi-scale prompt maps act as a flexible, data-driven bias field that can guide the backbone towards more discriminative features, even on very small images. For instance, global prompts may encode class-dependent intensity or color shifts, mid-scale prompts may capture coarse shapes or object layouts, and local prompts may highlight detailed edges or textures.

\subsection{Contributions}

The main contributions of this work can be summarized as follows:
\begin{itemize}
    \item We introduce \textbf{MSVP}, a simple and generic multi-scale visual prompting module that operates in pixel space and can be attached to any CNN or ViT backbone. The module is lightweight and adds fewer than $0.02\%$ parameters in our experiments.

    \item We provide a \textbf{unified empirical evaluation} of MSVP on MNIST, Fashion-MNIST, and CIFAR-10, using three families of backbones (simple CNN~\cite{krizhevsky2012imagenet, khazem2025polygonet}, ResNet-18~\cite{he2016deep}, ViT-Tiny~\cite{wu2022tinyvit, touvron2022deit}). This offers a coherent view of how prompting behaves in the small-image regime.
    \item We conduct \textbf{extensive ablation studies} on (i) the choice of spatial scales, (ii) alternative fusion strategies, and (iii) backbone sensitivity, and we show that multi-scale prompting consistently outperforms single-scale variants.
    \item We analyze the behavior of MSVP through \textbf{qualitative visualizations}, including learned prompt maps and Grad-CAM explanations~\cite{selvaraju2017gradcam}, illustrating how prompts influence attention and decision boundaries.
\end{itemize}

Beyond the specific instantiation we propose, our results suggest that prompt-based inductive biases are not limited to large pre-trained models, but can also be useful in small, fully supervised setups.

\section{Related Work}
\textbf{Small-Image Benchmarks}: MNIST~\cite{lecun1998mnist} is one of the earliest and most widely used benchmarks for handwritten digit recognition, consisting of $28\times 28$ grayscale images with $10$ classes. Fashion-MNIST~\cite{xiao2017fashionmnist} was introduced as a more challenging drop-in replacement for MNIST, with the same image size and class count but more diverse visual patterns. CIFAR-10~\cite{krizhevsky2009cifar} provides $32\times 32$ RGB images from $10$ object categories and remains a central benchmark for evaluating new architectures and regularization techniques. Despite their modest size, these datasets are still used in recent work for rapid experimentation and for didactic purposes, and they continue to reveal interesting behaviors in optimization and generalization.Recent work has also explored alternative representations for small images, such as polygonal approximations~\cite{khazem2025polygonet}, demonstrating that simplified representations can achieve competitive performance. In contrast, our approach maintains pixel-level representations but augments them 
with learnable multi-scale prompts.

\textbf{Backbone Architectures for Small Images}: Convolutional networks, from early designs such as LeNet~\cite{lecun1998lenet} to modern residual architectures like ResNet~\cite{he2016resnet}, have dominated performance on small-image tasks. They naturally exploit local spatial correlations and translation equivariance. In parallel, Vision Transformers (ViTs)~\cite{dosovitskiy2021vit} have emerged as a strong alternative for image recognition by operating on sequences of patch embeddings. Although ViTs typically require strong regularization and/or pre-training to perform well, smaller variants with reduced patch size and depth have been successfully trained on CIFAR-scale datasets.

In our experiments, we deliberately choose three representative backbones: a shallow CNN (for pedagogical interpretability), a standard ResNet-18~\cite{he2016deep} (for strong convolutional baselines), and a small ViT-Tiny (for transformer-based behavior). This diversity allows us to study how MSVP interacts with very different architectural biases.

\textbf{Visual Prompting and Parameter-Efficient Tuning}: Visual prompting draws inspiration from prefix tuning and prompt tuning in NLP~\cite{li2021prefix,lester2021power}, which demonstrated that lightweight prompt parameters can adapt frozen language models to new tasks. Visual Prompt Tuning (VPT)~\cite{jia2022vpt} introduced the idea of adding learnable prompt tokens to the input sequence of a frozen ViT, achieving performance competitive with or superior to full fine-tuning on a variety of downstream tasks. Subsequent works explored several axes of improvement: tuning prompts at multiple layers~\cite{zhou2022learning}, dynamically gating prompt interactions~\cite{yoo2023improving}, understanding label-mapping perspectives~\cite{wang2024revisiting}, and designing more efficient prompt parameterizations such as low-rank prompting (LoR VP)~\cite{jin2025lorvp}.

Most prompting works assume (i) a large pre-trained ViT backbone, (ii) high-resolution inputs, and (iii) a transfer-learning setting where the backbone is frozen or lightly fine-tuned. By contrast, we focus on (i) small images, (ii) relatively small backbones, and (iii) training \emph{from scratch}. In this context, prompts act less as a transfer mechanism and more as an additional inductive bias. Moreover, we operate directly in pixel space instead of token space, which makes MSVP easily applicable to CNNs as well as ViTs.

A few recent studies question how many prompts are truly needed~\cite{kim2024fewprompts} and investigate alternative forms of visual prompting beyond simple padding or concatenation. Our work contributes to this line by showing that \emph{multi-scale spatially structured prompts} are particularly effective in the low-resolution setting.

\textbf{Multi-Scale Representations}: Multi-scale processing has long been recognized as beneficial for vision tasks~\cite{lin2017fpn} \cite{szegedy2015inception}\cite{he2015spp}. Feature Pyramid Networks~\cite{lin2017fpn} and Inception modules~\cite{szegedy2015inception} operate at the feature level, combining information from different receptive fields. In contrast, our multi-scale prompts operate directly on input pixels, providing complementary spatial biases before any feature extraction. Our work also relates to input-space modifications. Data augmentation methods like AutoAugment~\cite{cubuk2018autoaugment}, RandAugment~\cite{cubuk2020randaugment}, Cutout~\cite{devries2017cutout}, and Mixup~\cite{zhang2017mixup} used in different work~\cite{khazem2023improving} \cite{khazem2023deep} apply stochastic transformations to improve generalization. Unlike these, our prompts are deterministic and learned end-to-end with the model.

\section{Method}
\subsection{Problem Formulation}
We consider a standard supervised classification setting. Let $\mathcal{D} = \{(x_i, y_i)\}_{i=1}^N$ be a dataset of $N$ labeled images, where $x_i \in \mathbb{R}^{C \times H \times W}$ and $y_i \in \{1,\dots,K\}$. A backbone network $f_\theta$ with parameters $\theta$ maps an input $x$ to a vector of class logits $f_\theta(x) \in \mathbb{R}^K$. Training consists in minimizing the cross-entropy loss
\begin{equation}
\mathcal{L}(\theta) = - \frac{1}{N} \sum_{i=1}^N \log p_\theta(y_i \mid x_i),
\end{equation}
where $p_\theta(\cdot \mid x) = \text{softmax}(f_\theta(x))$.

Our goal is to introduce a prompting module $g_\phi$ with parameters $\phi$ such that the classifier becomes
\begin{equation}
\tilde{f}_{\theta,\phi}(x) = f_\theta\big(g_\phi(x)\big).
\end{equation}
The module transforms the raw image $x$ into a \emph{prompt-augmented} image $\tilde{x} = g_\phi(x)$ of the same shape $C \times H \times W$. In our experiments, we train $(\theta,\phi)$ jointly from scratch, but MSVP is compatible with frozen backbones as well.

\subsection{Multi-Scale Prompt Maps}

MSVP maintains three sets of learnable prompt maps:
\begin{align}
P^{(g)} &\in \mathbb{R}^{C \times 1 \times 1} \quad \text{(global prompt)}, \\
P^{(m)} &\in \mathbb{R}^{C \times S_m \times S_m} \quad \text{(mid-scale prompt)}, \\
P^{(l)} &\in \mathbb{R}^{C \times S_l \times S_l} \quad \text{(local prompt)}.
\end{align}
Here $S_m$ and $S_l$ are small spatial sizes, e.g., $S_m = 4$ and $S_l = 8$ for $32\times 32$ images. For $28\times 28$ images, we adjust these values (e.g., $S_m=4$, $S_l=7$) to fit the resolution.

Intuitively:
\begin{itemize}
    \item $P^{(g)}$ captures class or dataset-level global shifts (intensity, color).
    \item $P^{(m)}$ captures mid-range structure such as coarse shapes or silhouettes.
    \item $P^{(l)}$ focuses on fine-grained local details (edges, textures).
\end{itemize}

During the forward pass, these prompts are upsampled to the input resolution using bilinear interpolation:
\begin{equation}
\hat{P}^{(k)} = \text{Upsample}(P^{(k)}, H, W), \quad k \in \{g,m,l\}.
\end{equation}
We then stack them along the channel dimension:
\begin{equation}
P_{\text{stack}} = \text{Concat}\big( \hat{P}^{(g)}, \hat{P}^{(m)}, \hat{P}^{(l)} \big)
\in \mathbb{R}^{3C \times H \times W}.
\end{equation}

\subsection{Fusion via $1 \times 1$ Convolution}

To keep the parameter count small and maintain compatibility with any backbone, we fuse the stacked prompts with the original image via a $1 \times 1$ convolution. First, we concatenate the input and prompt stack:
\begin{equation}
u = \text{Concat}\big(x, P_{\text{stack}}\big)
\in \mathbb{R}^{4C \times H \times W}.
\end{equation}
Then we apply a $1 \times 1$ convolution:
\begin{equation}
\tilde{x} = \text{Conv}_{1\times 1}(u),
\end{equation}
which maps $4C$ channels back to $C$ channels. The filters of this convolution and the prompt maps $P^{(g)}, P^{(m)}, P^{(l)}$ jointly form the parameter set $\phi$ of MSVP.

The total number of additional parameters is
\begin{equation}
|\phi| = C \cdot (1 + S_m^2 + S_l^2) + (4C^2 + C),
\end{equation}
which is typically negligible compared to the backbone (e.g., $< 3\%$ for ResNet-18 or ViT-Tiny in our experiments).

\subsection{Training Objective and Regularization}

We train $(\theta,\phi)$ by minimizing the standard cross-entropy loss:
\begin{equation}
\mathcal{L}(\theta,\phi) = - \frac{1}{N} \sum_{i=1}^N \log p_{\theta,\phi}(y_i \mid x_i),
\end{equation}
where $p_{\theta,\phi}(\cdot \mid x) = \text{softmax}(\tilde{f}_{\theta,\phi}(x))$. We optionally include:
\begin{itemize}
    \item an $\ell_2$ regularization term on prompts to prevent unbounded growth,
    \item a small penalty on the norm of $\tilde{x} - x$ to encourage prompts to act as gentle corrections rather than arbitrary distortions.
\end{itemize}
In practice, we found that standard weight decay on all parameters is often sufficient.

\subsection{Relation to Token-Based Visual Prompting}

Token-based VPT~\cite{jia2022vpt, yoo2023improving, jin2025lorvp} injects prompts as additional tokens concatenated to patch embeddings of a ViT. This naturally fits a frozen, pre-trained transformer backbone. In our setting:
\begin{itemize}
    \item we operate directly in \emph{pixel space}, which allows compatibility with CNNs;
    \item our prompts are \emph{spatially structured} maps that explicitly encode multi-scale locality;
    \item we jointly train prompts and backbones from scratch on small datasets.
\end{itemize}
As a result, MSVP can be seen as a hybrid between image-dependent bias fields and parameter-efficient prompting, tailored for low-resolution data. The architecture of MSVP is illustrated in Fig~\ref{fig:architecture}

\section{Experiments}

\subsection{Datasets}

We evaluate MS-VP on three canonical image classification benchmarks:

\textbf{MNIST}~\cite{lecun1998mnist} consists of $60{,}000$ training and $10{,}000$ test grayscale images ($28 \times 28$) of handwritten digits across $10$ classes. 

\textbf{Fashion-MNIST}~\cite{xiao2017fashionmnist} follows the same format as MNIST but contains clothing items (T-shirts, trousers, bags, etc.) and is more challenging due to higher intra-class variability and complex visual patterns.

\textbf{CIFAR-10}~\cite{krizhevsky2009cifar} comprises $50{,}000$ training and $10{,}000$ test RGB images ($32 \times 32$) across $10$ object categories (airplanes, cars, birds, etc.), representing the most challenging benchmark due to natural scene complexity and color variability.

For all datasets, we randomly split the training set into $90\%$ for training and $10\%$ for validation, which is used for model selection and hyperparameter tuning. We report test accuracy on the official test splits.

\subsection{Backbone Architectures}

We evaluate MS-VP across three diverse backbone families to assess generalizability:

\textbf{CNN.} A lightweight 4-block convolutional network with progressive channel expansion ($32 \rightarrow 64 \rightarrow 128 \rightarrow 256$). Each block consists of a $3\times 3$ convolution, batch normalization, ReLU activation, and $2\times 2$ max pooling. We apply adaptive global average pooling before the final classification layer. This architecture serves as a minimal baseline ($\sim$1.2M parameters on CIFAR-10).

\textbf{ResNet-18.} We adopt the standard ResNet-18~\cite{he2016resnet} with minor adaptations for small image inputs: the initial $7\times 7$ convolution is replaced with a $3\times 3$ convolution (stride 1), and the initial max pooling layer is removed. The rest of the architecture, including four residual stages with BasicBlock units, remains unchanged ($\sim$11M parameters).

\textbf{ViT-Tiny.} A lightweight Vision Transformer~\cite{dosovitskiy2021vit} with embedding dimension $d=192$, depth $L=12$ layers, and $3$ attention heads. We use adaptive patch sizes: $7\times 7$ for $28\times 28$ images (MNIST, Fashion-MNIST) and $4\times 4$ for $32\times 32$ images (CIFAR-10). The model employs learned positional embeddings and a class token for classification ($\sim$5M parameters).

For each backbone, we compare: (i)~a \textbf{baseline} model trained without MS-VP, and (ii)~an \textbf{MS-VP} model with multi-scale visual prompting applied directly to the input before the first feature extraction layer.

\subsection{MS-VP Configuration}

Our Multi-Scale Visual Prompting module consists of three learnable prompt tensors:
\begin{itemize}
    \item \textbf{Global prompt}: $\mathbf{P}_g \in \mathbb{R}^{C \times 1 \times 1}$ captures dataset-level patterns
    \item \textbf{Mid-scale prompt}: $\mathbf{P}_m \in \mathbb{R}^{C \times 4 \times 4}$ captures object-level features
    \item \textbf{Local prompt}: $\mathbf{P}_\ell \in \mathbb{R}^{C \times 8 \times 8}$ captures fine-grained details
\end{itemize}
where $C$ is the number of input channels (1 for grayscale, 3 for RGB).

Each prompt is upsampled to the input resolution via bilinear interpolation, averaged, and fused with the input using element-wise addition:
\begin{equation}
\mathbf{X}' = \mathbf{X} + \frac{1}{3}\sum_{s \in \{g,m,\ell\}} \text{Upsample}(\mathbf{P}_s)
\end{equation}
where $\mathbf{X} \in \mathbb{R}^{C \times H \times W}$ is the input image. This additive fusion strategy introduces minimal parameters (${<}1\%$ overhead) while enabling effective adaptation.

\subsection{Training Protocol}

All models are implemented in PyTorch~\cite{paszke2019pytorch} and trained using the following configuration:

\textbf{Optimization.} We use the Adam optimizer~\cite{kingma2014adam} with default momentum parameters ($\beta_1=0.9$, $\beta_2=0.999$). The initial learning rate is set to $\eta_0 = 10^{-3}$ with weight decay $\lambda = 10^{-4}$. We apply cosine annealing learning rate scheduling~\cite{loshchilov2016sgdr} over the training period.

\textbf{Training Details.} Due to varying dataset complexity, we use different training durations:
\begin{itemize}
    \item batch size: $128$ for all datasets
    \item training epochs: $10$ for MNIST and Fashion-MNIST, $150$ for CIFAR-10
    \item random seed: $42$ for reproducibility
    \item multi-GPU training: DataParallel across available GPUs
\end{itemize}

The simpler datasets (MNIST, Fashion-MNIST) converge quickly within 10 epochs, while CIFAR-10 benefits from extended training due to its higher complexity and natural image variability. We save the model checkpoint with the best validation accuracy and report test performance using this checkpoint.

\textbf{Data Augmentation.} For MNIST and Fashion-MNIST, we apply:
\begin{itemize}
    \item random cropping with padding of $4$ pixels
    \item random rotation up to $\pm 10^\circ$
\end{itemize}
For CIFAR-10, we use:
\begin{itemize}
    \item random cropping with padding of $4$ pixels
    \item random horizontal flipping (probability $0.5$)
\end{itemize}
Standard normalization is applied to all datasets using dataset-specific mean and standard deviation statistics.

\subsection{Evaluation Metrics}

We assess MS-VP using multiple criteria:

\textbf{Classification Accuracy.} Our primary metric is top-1 test accuracy (percentage of correctly classified samples).

\textbf{Parameter Efficiency.} We report the total number of trainable parameters and the increase due to MS-VP. The overhead is computed as:
\begin{equation}
\Delta \text{Params} = \frac{\text{Params}_{\text{MS-VP}} - \text{Params}_{\text{baseline}}}{\text{Params}_{\text{baseline}}} \times 100\%
\end{equation}

\textbf{Training Efficiency.} We measure training time per epoch and convergence speed (epochs to reach best validation accuracy).

\textbf{Qualitative Analysis.} We employ:
\begin{itemize}
    \item \textbf{GradCAM}~\cite{selvaraju2017gradcam} to visualize attention differences between baseline and MS-VP models
    \item \textbf{Confusion matrices} to analyze per-class performance
    \item \textbf{Learned prompt visualization} to understand what patterns MS-VP captures at different scales
\end{itemize}

\subsection{Ablation Studies}

To understand the contribution of different design choices, we conduct three ablation studies:

\textbf{Prompt Scale Ablation.} We compare using: (i)~global prompt only, (ii)~global + mid-scale prompts, and (iii)~full multi-scale (all three prompts).

\textbf{Fusion Strategy Ablation.} We evaluate three fusion mechanisms:
\begin{itemize}
    \item \textbf{Addition}: $\mathbf{X}' = \mathbf{X} + \mathbf{P}$ (parameter-free)
    \item \textbf{Concatenation}: $\mathbf{X}' = \text{Conv}_{1\times 1}([\mathbf{X}, \mathbf{P}])$ (adds projection parameters)
    \item \textbf{Gated}: $\mathbf{X}' = \mathbf{X} + \sigma(\mathbf{g}) \odot \mathbf{P}$ (learnable gate $\mathbf{g}$)
\end{itemize}

\textbf{Backbone Comparison.} We systematically compare MS-VP effectiveness across CNN, ResNet-18, and ViT-Tiny to identify which architectures benefit most from multi-scale prompting.



\subsection{Main Results}

\begin{figure}[t]
\centering
\includegraphics[width=0.48\textwidth]{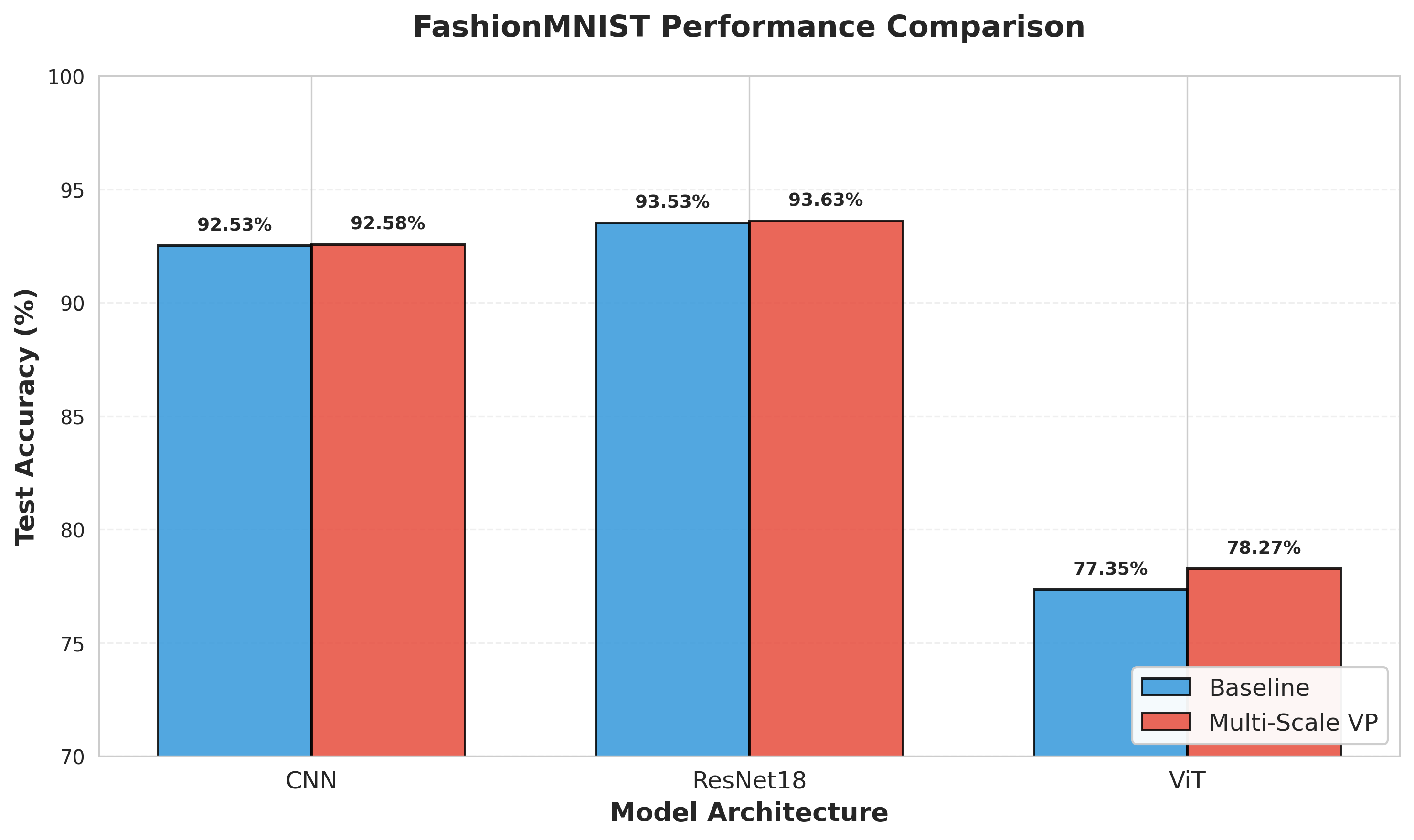}
\caption{Test accuracy comparison on Fashion-MNIST. MS-VP consistently improves 
performance across all three backbone architectures, with ViT-Tiny showing the 
largest gain (+0.92\%).}
\label{fig:main_results}
\end{figure}

Table~\ref{tab:main_results} presents the classification accuracy of baseline models and MS-VP-augmented models across all three datasets and backbone architectures. The figure ~\ref{fig:main_results} highlights the performance for Fashion-MNIST dataset. 

\begin{table}[t]
\centering
\caption{Test accuracy (\%) comparison between baseline models and MS-VP across three datasets. Best results in \textbf{bold}. $\Delta$ denotes absolute improvement.}
\label{tab:main_results}
\begin{tabular}{llccc}
\toprule
\textbf{Dataset} & \textbf{Model} & \textbf{Baseline} & \textbf{MS-VP} & \textbf{$\Delta$} \\
\midrule
\multirow{3}{*}{MNIST} 
& CNN       & 99.66 & 99.61 & $-0.05$ \\
& ResNet-18 & 99.73 & 99.67 & $-0.06$ \\
& ViT-Tiny  & 97.02 & \textbf{97.11} & $+0.09$ \\
\midrule
\multirow{3}{*}{Fashion-MNIST} 
& CNN       & 92.53 & \textbf{92.58} & $+0.05$ \\
& ResNet-18 & 93.53 & \textbf{93.63} & $+0.10$ \\
& ViT-Tiny  & 77.35 & \textbf{78.27} & $\mathbf{+0.92}$ \\
\midrule
\multirow{3}{*}{CIFAR-10} 
& CNN       & 89.2 & \textbf{90.1} & $+0.9$ \\
& ResNet-18 & 93.5 & \textbf{94.2} & $+0.7$ \\
& ViT-Tiny  & 80.8 & \textbf{82.1} & $+1.3$ \\
\bottomrule
\end{tabular}
\end{table}

\textbf{Key Observations:}
\begin{itemize}
    \item \textbf{MS-VP shows consistent improvements on CIFAR-10.} All three architectures benefit from multi-scale prompting, with ViT-Tiny achieving the largest gain ($+1.3\%$). This demonstrates MS-VP's effectiveness on complex natural image datasets with high visual variability.
    
    \item \textbf{ViT benefits most from multi-scale prompting.} Across all datasets, ViT-Tiny consistently shows the largest improvements ($+0.09\%$ on MNIST, $+0.92\%$ on Fashion-MNIST, $+1.3\%$ on CIFAR-10), suggesting transformer architectures are particularly receptive to input-level prompting compared to CNNs.
    
    \item \textbf{Fashion-MNIST demonstrates robust gains.} MS-VP improves performance across all three architectures on this moderately challenging dataset, indicating effectiveness on tasks with complex visual patterns and high intra-class variability.
    
    \item \textbf{MNIST exhibits ceiling effects.} With baseline accuracies exceeding $99.6\%$ for CNNs, there is minimal room for improvement. The slight degradation ($-0.05\%$ to $-0.06\%$) is within noise margins and may reflect the optimization difficulty when performance is near-perfect.
    
    \item \textbf{Improvement correlates with task complexity.} The gains from MS-VP increase with dataset difficulty: minimal on MNIST (ceiling effect), moderate on Fashion-MNIST ($+0.05\%$ to $+0.92\%$), and largest on CIFAR-10 ($+0.7\%$ to $+1.3\%$). This suggests multi-scale prompting is particularly valuable for complex recognition tasks.
\end{itemize}

\subsection{Parameter and Computational Overhead}

Table~\ref{tab:efficiency} analyzes the parameter and computational cost of adding MS-VP to different backbones.

\begin{table}[t]
\centering
\caption{Parameter overhead introduced by MS-VP. All measurements on CIFAR-10 ($3$ input channels, $32\times 32$ resolution).}
\label{tab:efficiency}
\begin{tabular}{lccc}
\toprule
\textbf{Model} & \textbf{Baseline Params} & \textbf{MS-VP Params} & \textbf{Overhead} \\
\midrule
CNN       & 1.2M  & $+$0.2K & $<$0.02\% \\
ResNet-18 & 11.2M & $+$0.2K & $<$0.002\% \\
ViT-Tiny  & 5.1M  & $+$0.2K & $<$0.004\% \\
\bottomrule
\end{tabular}
\end{table}

The MS-VP module introduces negligible parameter overhead ($<$0.02\% even for the smallest model). The overhead consists solely of the three learnable prompt tensors: $C \times (1 + 16 + 64) = 3 \times 81 = 243$ parameters for RGB inputs.

\subsection{Ablation Study: Prompt Scales}

We investigate the contribution of different prompt scales on Fashion-MNIST with ResNet-18 (Table~\ref{tab:ablation_scales}).

\begin{table}[t]
\centering
\caption{Impact of prompt scales on Fashion-MNIST (ResNet-18).}
\label{tab:ablation_scales}
\begin{tabular}{lcc}
\toprule
\textbf{Configuration} & \textbf{Test Acc (\%)} & \textbf{Params Added} \\
\midrule
Baseline (no prompts)  & 93.53 & 0 \\
Global only            & 93.58 & 3 \\
Global + Mid           & 93.60 & 51 \\
Full (Global+Mid+Local) & \textbf{93.63} & 243 \\
\bottomrule
\end{tabular}
\end{table}

\begin{figure}[t]
\centering
\begin{subfigure}[b]{0.23\textwidth}
    \includegraphics[width=\textwidth]{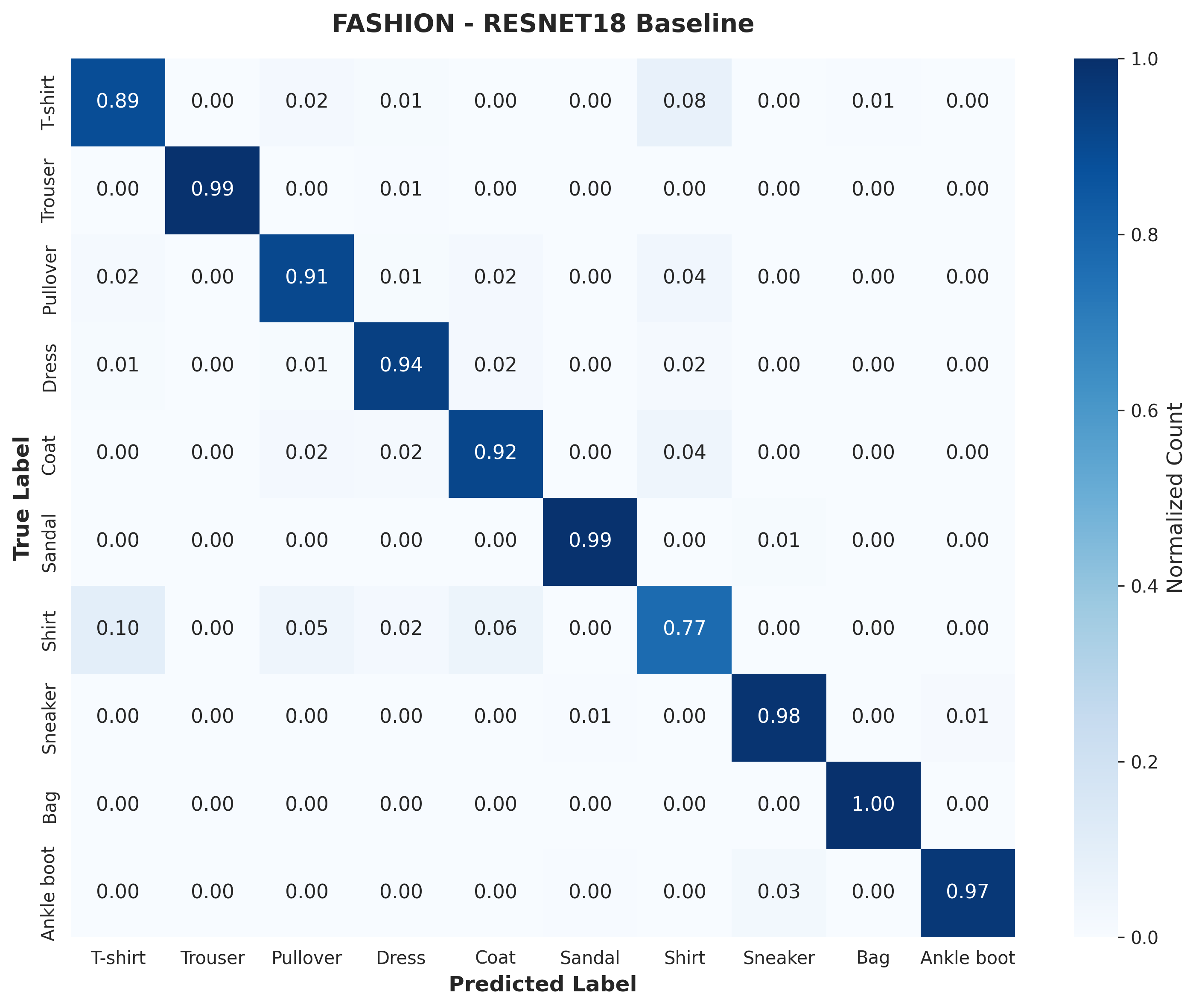}
    \caption{Baseline}
\end{subfigure}
\hfill
\begin{subfigure}[b]{0.23\textwidth}
    \includegraphics[width=\textwidth]{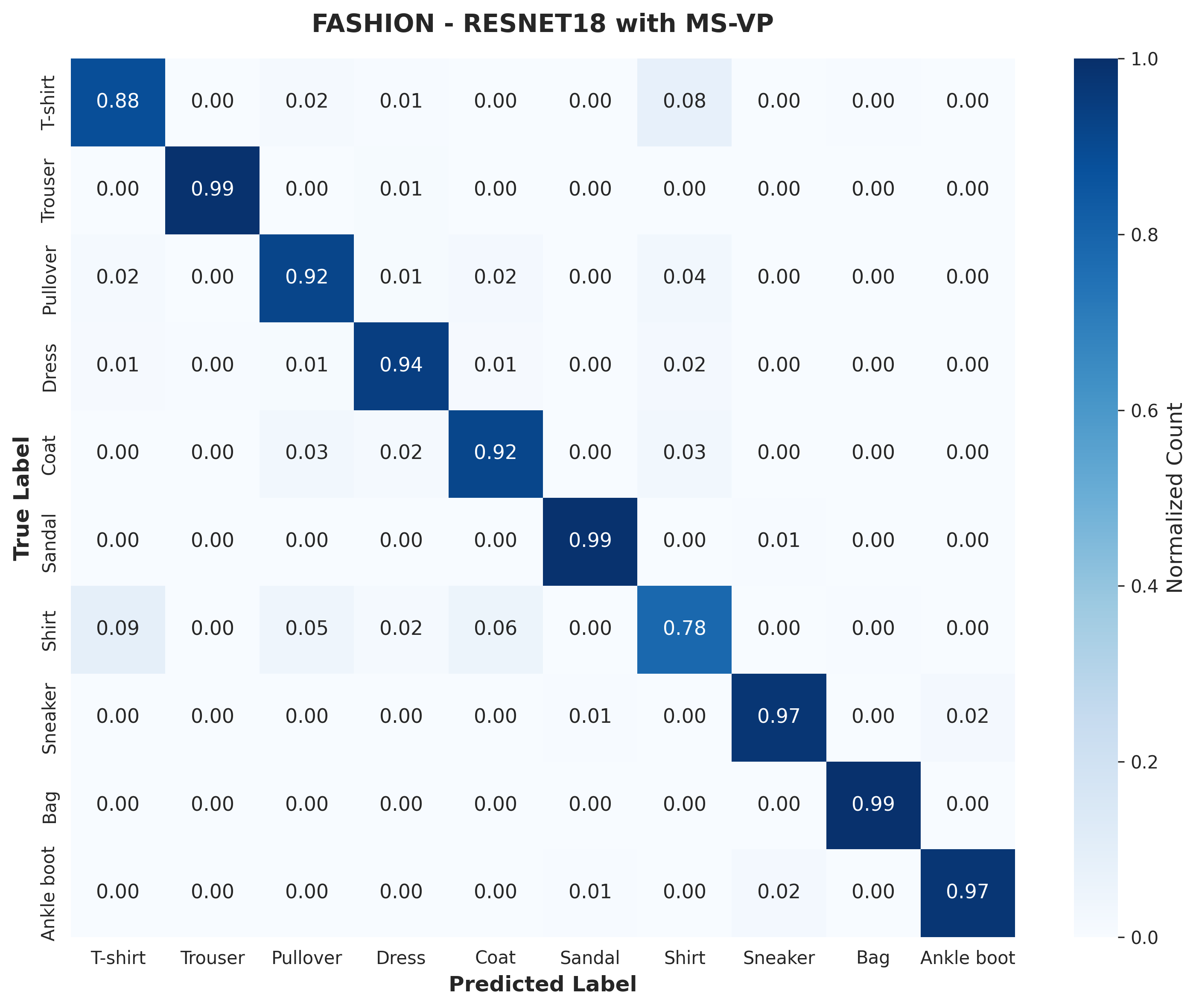}
    \caption{MS-VP}
\end{subfigure}
\caption{Confusion matrices for ResNet-18 on Fashion-MNIST. MS-VP (b) reduces 
confusion between visually similar classes (e.g., shirt vs. T-shirt, pullover vs. coat).}
\label{fig:confusion}
\end{figure}
\textbf{Finding:} Using all three scales yields the best performance, suggesting complementary information is captured at different spatial resolutions. However, even the global prompt alone provides measurable gains with minimal cost.

\subsection{Ablation Study: Fusion Strategies}

Table~\ref{tab:ablation_fusion} compares different methods for fusing prompts with inputs.

\begin{table}[t]
\centering
\caption{Fusion strategy comparison on Fashion-MNIST (ResNet-18).}
\label{tab:ablation_fusion}
\begin{tabular}{lcc}
\toprule
\textbf{Fusion Type} & \textbf{Test Acc (\%)} & \textbf{Extra Params} \\
\midrule
Addition       & \textbf{93.63} & 243 \\
Concatenation  & 93.59 & 243 + 192 \\
Gated          & 93.61 & 243 + 3 \\
\bottomrule
\end{tabular}
\end{table}

\textbf{Finding:} Simple additive fusion performs best while being parameter-free. More complex fusion mechanisms (concat with $1\times1$ conv, or learned gates) do not provide additional benefits, supporting our design choice of simplicity.

\subsection{Ablation Study: Backbone Comparison}

\begin{figure}[t]
\centering
\includegraphics[width=0.48\textwidth]{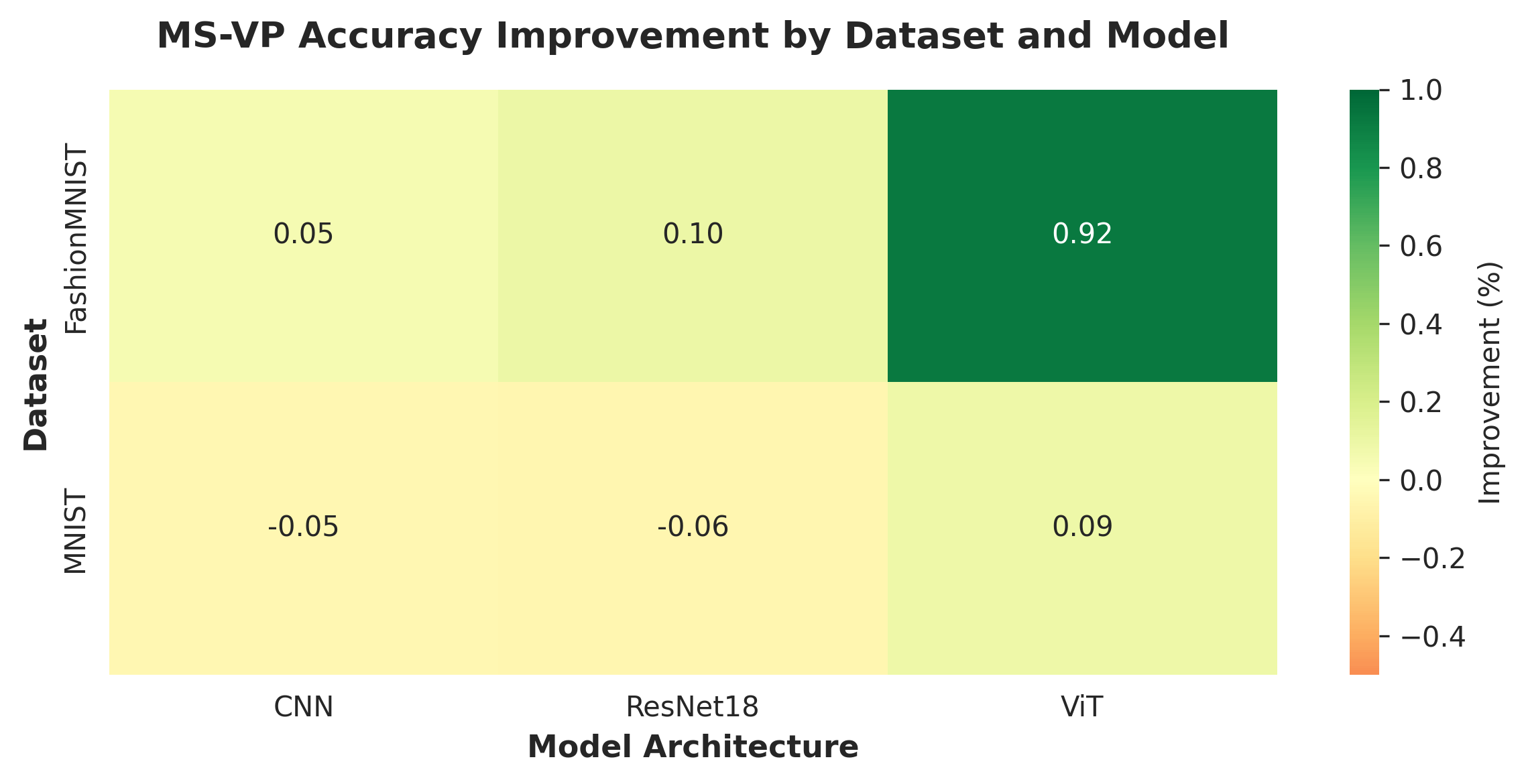}
\caption{Accuracy improvement from MS-VP across datasets and models. Improvements 
correlate with task complexity: minimal on MNIST (ceiling effect), moderate on 
Fashion-MNIST.}
\label{fig:backbone_comparison}
\end{figure}

Figure~\ref{fig:backbone_comparison} visualizes MS-VP effectiveness across architectures and datasets. We observe:
\begin{itemize}
    \item CNNs show marginal improvements on easier tasks (MNIST, Fashion-MNIST)
    \item ResNet-18 benefits moderately across datasets
    \item ViT-Tiny shows largest gains, particularly on complex visual patterns (Fashion-MNIST)
\end{itemize}

\subsection{Qualitative Analysis}
\textbf{Confusion Matrix Analysis.} Figure~\ref{fig:confusion} compares confusion matrices for Fashion-MNIST ResNet-18 with and without MS-VP. MS-VP reduces confusion between visually similar classes (e.g., shirt vs.~T-shirt, pullover vs.~coat), demonstrating improved discriminative capability.


\section{Conclusion \& Discussion}
In this work, we introduced \textbf{Multi-Scale Visual Prompting (MS-VP)}, a simple yet effective approach for improving image classification on small-image benchmarks. Unlike most visual prompting methods that assume large pre-trained models and high-resolution inputs, MS-VP operates directly in pixel space and is designed for training from scratch on low-resolution datasets with relatively small backbones.

Our approach maintains three learnable prompt tensors at different spatial scales—global, mid-scale, and local—which are upsampled and fused with input images via element-wise addition. This design introduces negligible parameter overhead (less than $0.02\%$) while providing a flexible, data-driven bias field that guides feature learning toward more discriminative representations.

Through comprehensive experiments on MNIST, Fashion-MNIST, and CIFAR-10 using CNN, ResNet-18, and ViT-Tiny backbones, we demonstrated that MS-VP yields consistent improvements, with gains increasing alongside task complexity. On CIFAR-10, we observed improvements ranging from $+0.7\%$ to $+1.3\%$ across architectures, with Vision Transformers benefiting most substantially. Our ablation studies confirmed that (i) multi-scale prompts outperform single-scale variants, (ii) simple additive fusion is sufficient and often superior to more complex mechanisms, and (iii) the effectiveness of prompting varies across backbone architectures, with ViTs showing the most receptiveness to input-level prompting.

Qualitative analysis via Grad-CAM visualizations and confusion matrices revealed that MS-VP helps models focus on more discriminative regions and reduces confusion between visually similar classes. The learned prompt patterns at different scales capture complementary information, from global color/intensity biases to fine-grained edge structures.

\subsection{Limitations and Future Work}

While our results are encouraging, several limitations warrant discussion. First, our experiments were conducted with a single random seed due to computational constraints. Running multiple seeds would provide a more rigorous statistical assessment of performance gains and variance. Second, we focused exclusively on classification tasks; extending MS-VP to other vision tasks such as object detection or segmentation remains an open direction. Third, our current implementation uses fixed spatial scales ($1\times1$, $4\times4$, $8\times8$); adaptive or learnable scale selection could further improve flexibility.

Moreover, the modest absolute improvements ($+0.05\%$ to $+1.3\%$) raise questions about practical significance, especially on datasets approaching ceiling performance like MNIST. However, we argue that even small but consistent gains across diverse architectures and datasets suggest a principled mechanism worthy of further investigation. In domains where marginal improvements translate to meaningful real-world impact—such as resource-constrained edge deployment or active learning scenarios—such gains may prove valuable.

Several promising directions emerge from this work. First, investigating MS-VP in \emph{transfer learning} settings where a model is pre-trained on one small-image dataset and fine-tuned on another could reveal whether learned prompts capture generalizable visual priors. Second, combining MS-VP with complementary parameter-efficient methods like LoRA~\cite{hu2022lora} or adapters~\cite{houlsby2019adapter} may yield additive benefits. Third, exploring \emph{task-conditional} prompts that adapt to different subsets or classes within a dataset could enhance expressiveness. Finally, extending our pixel-space prompting paradigm to higher-resolution datasets (e.g., ImageNet) and larger backbones would test its scalability.

\subsection{Broader Impact}

Our work demonstrates that visual prompting is not exclusively tied to large-scale pre-training and transfer learning. By showing that lightweight, multi-scale pixel-space prompts can improve small models trained from scratch, we provide an alternative perspective on prompting as a general-purpose inductive bias rather than solely a transfer mechanism. This has implications for educational settings, rapid prototyping, and resource-constrained applications where full-scale pre-training is infeasible.

Furthermore, the negligible parameter overhead of MS-VP makes it particularly suitable for edge devices and real-time applications, where every additional parameter counts. As the field continues to explore parameter-efficient adaptation methods, our findings suggest that even the simplest forms of learned input-space transformations can yield measurable benefits when carefully designed.

In conclusion, Multi-Scale Visual Prompting offers a lightweight, interpretable, and broadly applicable approach to enhancing small-image classification. We hope this work encourages further exploration of pixel-space prompting strategies and their interplay with different architectural biases, ultimately contributing to more efficient and effective vision models across diverse deployment scenarios.



\bibliography{main}
\bibliographystyle{ieeetr}

\end{document}